%% file: neurips_2026.tex
\documentclass{article}

\PassOptionsToPackage{numbers,sort&compress}{natbib}

 \usepackage[preprint]{neurips_2026}

\usepackage[utf8]{inputenc} 
\usepackage[T1]{fontenc}    
\usepackage{hyperref}       
\usepackage{url}            
\usepackage{booktabs}       
\usepackage{amsfonts}       
\usepackage{nicefrac}       
\usepackage{microtype}      
\usepackage{xcolor}         
\usepackage{amsmath}
\usepackage[table]{xcolor}
\usepackage[normalem]{ulem}
\definecolor{darkgreen}{RGB}{0,100,0}
\definecolor{darkred}{RGB}{139,0,0}
\usepackage{makecell}
\usepackage{array}
\usepackage{multirow}
\usepackage{threeparttable}

\usepackage{graphicx}
\usepackage{wrapfig}
\let\origincludegraphics\includegraphics
\renewcommand{\includegraphics}[2][]{%
  \IfFileExists{#2}{%
    \origincludegraphics[#1]{#2}%
  }{%
    \fbox{\parbox[c][0.18\textheight][c]{0.9\linewidth}{\centering Missing figure: \texttt{#2}}}%
  }%
}

\title{EmbodiedClaw: Conversational Workflow Execution for Embodied AI Development}

\author{%
  \parbox{\textwidth}{\centering
  {\bfseries Xueyang Zhou$^{1}$, Yihan Sun$^{2}$, Xijie Gong$^{3}$, Guiyao Tie$^{1}$}\\
  {\bfseries Pan Zhou$^{1}$, Lichao Sun$^{4}$, Yongchao Chen$^{5}$}\\[0.4em]
  {\normalfont $^{1}$School of Cyber Science and Engineering, Huazhong University of Science and Technology\\
  $^{2}$School of Mechanical Science and Engineering, Huazhong University of Science and Technology\\
  $^{3}$School of Information and Software Engineering, University of Electronic Science and Technology of China\\
  $^{4}$Lehigh University\\
  $^{5}$College of AI, Tsinghua University}\\[0.4em]
  {\normalfont \texttt{d202480819@hust.edu.cn}, \texttt{panzhou@hust.edu.cn}, \texttt{yongchaochen12@gmail.com}}}
}

\begin{document}

\maketitle

\begin{abstract}
  Embodied AI research is increasingly moving beyond single-task, single-environment policy learning toward multi-task, multi-scene, and multi-model settings. This shift substantially increases the engineering overhead and development time required for stages such as evaluation environment construction, trajectory collection, model training, and evaluation. To address this challenge, we propose a new paradigm for embodied AI development in which users express goals and constraints through conversation, and the system automatically plans and executes the development workflow. We instantiate this paradigm with EmbodiedClaw, a conversational agent that turns high-frequency, high-cost embodied research activities, including environment creation and revision, benchmark transformation, trajectory synthesis, model evaluation, and asset expansion, into executable skills. Experiments on end-to-end workflow tasks, capability-specific evaluations, human researcher studies, and ablations show that EmbodiedClaw reduces manual engineering effort while improving executability, consistency, and reproducibility. These results suggest a shift from manual toolchains to conversationally executable workflows for embodied AI development.
\end{abstract}

\input{sections/introduction}
\input{sections/preliminary}
\input{sections/method}
\input{sections/experiments}
\input{sections/related_works}
\input{sections/conclusion}

{\small\bibliographystyle{unsrtnat}\bibliography{references}}

\end{document}

%% file: sections/introduction.tex
\section{Introduction}
As embodied AI moves from single-task, single-environment research prototypes toward multi-task, multi-scene, and multi-model development, the bottleneck of embodied research is shifting from solving individual tasks to managing increasingly complex development workflows~\cite{duan2024survey,liu2023survey,feng2025embodied_survey,shao2025large,liu2025embodied,nahavandi2024machine}. Researchers now spend substantial time on environment construction, trajectory collection, model training, and evaluation, while repeatedly adapting tasks, assets, and model backends across platforms. This engineering overhead slows the iteration of ideas and algorithms, increases development cost, and makes embodied research harder to execute, reproduce, and extend~\cite{awais2025foundation,ma2025ecp,mu2025robotwin,hua2024gensim2,wang2023robogen}.

Despite rapid progress in simulators, benchmarks, and automated data generation methods~\cite{kolve2017ai2thor,savva2019habitat,szot2021habitat2,xiang2020sapien,makoviychuk2021isaac,zhu2020robosuite,tao2024maniskill3,wang2023robogen,deitke2022procthor,yang2024holodeck}, embodied AI development remains highly platform-bound and difficult to scale. Much of this burden comes from learning platform-specific task definitions and asset organizations, deploying environments, and adapting code across simulators, benchmarks, and model stacks~\cite{li2023igibson,gan2020threedworld,nasiriany2024robocasa,dalal2023imitating,mandlekar2023mimicgen,gu2023maniskill2}. In practice, these efforts repeatedly concentrate in some high-frequency activities: simulation task creation, task revision, trajectory collection, model training and evaluation~\cite{james2020rlbench,li2023behavior,tao2024maniskill3}. Inspired by "vibe coding", we argue that embodied AI development should move from manually assembling platform-specific experimental loops to conversationally driven execution of embodied development skills.

Therefore, we present EmbodiedClaw, which turns high-frequency, time-consuming embodied engineering tasks into an executable workflow driven by user intent through dialogue. It organizes embodied development around three classes of objects: simulation environments, data, and models. On top of these objects, EmbodiedClaw supports three core capabilities: simulation environment synthesis, trajectory data collection, and model deployment. This formulation is motivated by recent progress in language-driven embodied agents and general-purpose workflow automation systems~\cite{claudecode2026,codex2026,geminicli2026,openclaw2026}, while targeting the more structured and tool-dependent setting of embodied AI development~\cite{huang2022language,liang2022code,wang2023voyager,driess2023palme,zitkovich2023rt2}.

We evaluate this paradigm through end-to-end workflow tasks, capability-specific evaluations, human researcher studies, and ablations. Our results show that EmbodiedClaw reduces manual engineering burden while improving executability, consistency, and reproducibility in embodied AI development.

Our main contributions are as follows:
\begin{itemize}
    \item We propose a new view of embodied AI research in which high-frequency and time-intensive engineering activities are unified as a class of conversationally driven executable development skills. This abstraction turns previously fragmented engineering routines into automatic workflows and compresses workflows that previously required days of manual effort into only a few hours.
    \item We introduce EmbodiedClaw, a dialogue-driven agent system that translates user input into executable workflows and supports arbitrary operations over three core embodied research objects: simulation environments, data, and models.
    \item Extensive experiments show that EmbodiedClaw delivers substantial gains on a range of high-cost embodied development tasks, including batch simulation environment creation and editing, trajectory data collection, and model deployment and evaluation. In these settings, EmbodiedClaw achieves up to an order-of-magnitude improvement in efficiency while reaching accuracy close to that of human experts.
\end{itemize}

%% file: sections/preliminary.tex
\section{Preliminary}

\begin{figure}[t]
    \centering
    \includegraphics[width=\linewidth]{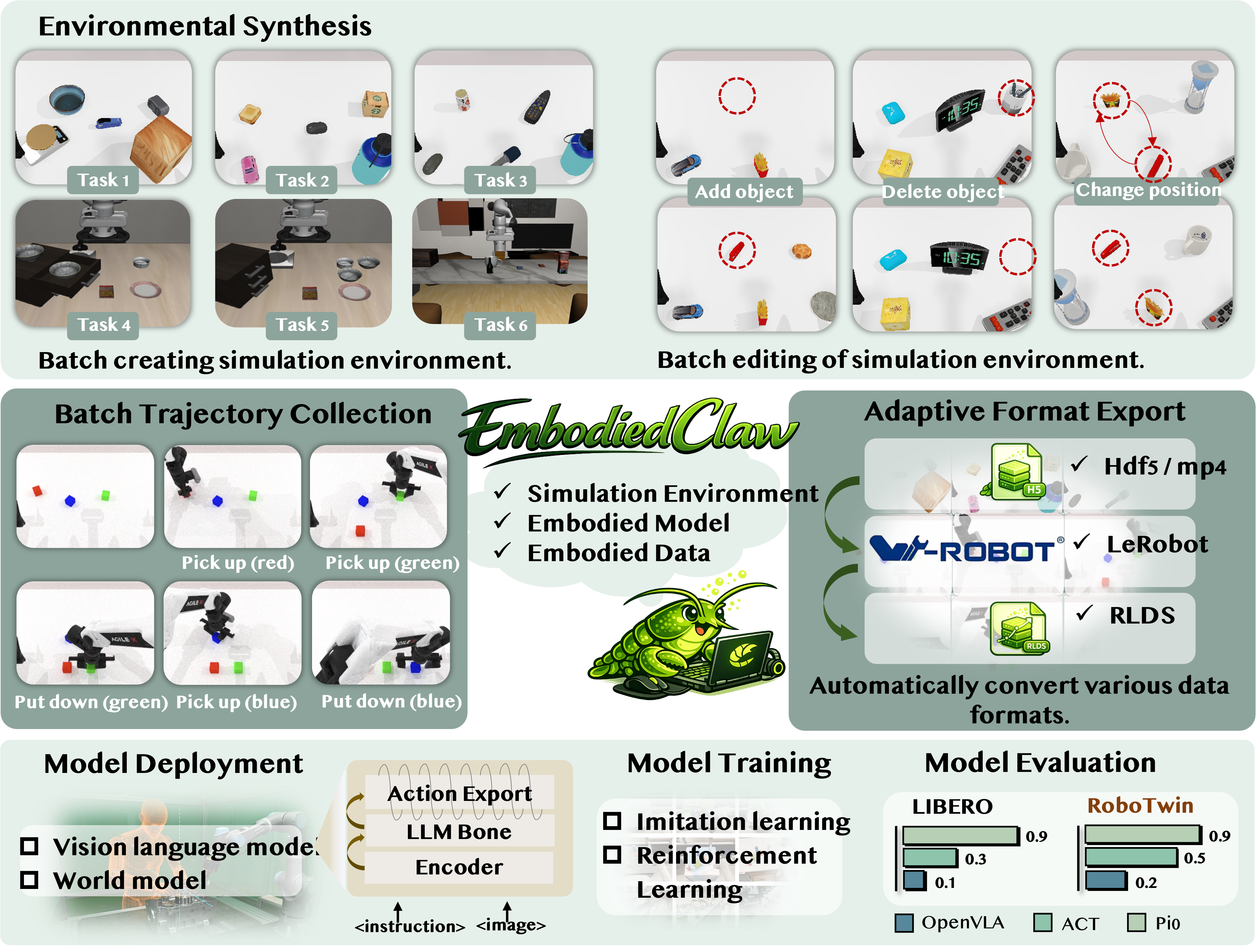}
    \caption{\textbf{Overview of EmbodiedClaw capabilities.} The framework supports batch simulation environment synthesis, including automatic scene creation and controllable scene editing, batch trajectory collection for embodied manipulation tasks, and adaptive export of embodied data to HDF5, LeRobot, RLDS, and video formats. It further enables model deployment with vision-language and world models, model training through imitation learning and reinforcement learning, and model evaluation on standard benchmarks such as LIBERO~\cite{liu2023libero}, RoboTwin~\cite{mu2025robotwin}, and SimplerEnv~\cite{li2024evaluating}.}
    \label{fig:accuracy_comparison}
\end{figure}

\subsection{EmbodiedClaw}
We formulate EmbodiedClaw as a conversational framework that maps user interaction into executable embodied development workflows over three operational objects: executable environments $o_t^{S}$, data $o_t^{D}$, and model/evaluation states $o_t^{M}$. At step $t$, the operational context is
\begin{equation}
\mathbf{o}_t = (o_t^{S}, o_t^{D}, o_t^{M}) \in \mathcal{O}^{S} \times \mathcal{O}^{D} \times \mathcal{O}^{M}.
\end{equation}
Given dialogue context $c_t$ and user input $x_t$, EmbodiedClaw infers an intent representation $z_t^{I}=\psi(c_t,x_t)$ and uses it to synthesize an executable workflow $\pi_t = \phi(z_t^{I},\mathbf{o}_t)$, where $\pi_t \in \Pi$ is a sequence of skill calls. Conditioned on $z_t^{I}$, the workflow updates the operational objects according to
\begin{equation}
(o_t^{S}, o_t^{D}, o_t^{M}) \xrightarrow{\pi_t \mid z_t^{I}} (o_{t+1}^{S}, o_{t+1}^{D}, o_{t+1}^{M}).
\end{equation}
The objective is to minimize development cost while producing outcomes aligned with the inferred intent. We therefore define
\begin{equation}
\pi_t^{*}
=
\arg\min_{\pi \in \Pi}
\underbrace{\mathcal{C}_{\mathrm{human}}(\pi)+\alpha\,\mathcal{C}_{\mathrm{sys}}(\pi)}_{\text{Cost / Overhead}}
+
\underbrace{\lambda \, d\!\left((o_{t+1}^{S}, o_{t+1}^{D}, o_{t+1}^{M}),\mathbf{o}^{*}_{\mathrm{op}}(z_t^{I})\right)}_{\text{Goal Deviation}},
\end{equation}
where $\mathcal{C}_{\mathrm{human}}(\pi)$ denotes the required human effort, $\mathcal{C}_{\mathrm{sys}}(\pi)$ denotes system execution cost, including time and token usage, and $\mathbf{o}^{*}_{\mathrm{op}}(z_t^{I})$ denotes the target operational outcome implied by the inferred intent representation.

\subsection{Embodied AI Research Operations}
We identify three major classes of high-frequency and time-intensive operations in embodied AI research: simulation environment synthesis, trajectory synthesis, and model deployment. Given dialogue context $c_t$, user input $x_t$, and an optional initial simulator state $o_t^{S,\mathrm{sim}}$, the system infers an intent representation $z_t^{I}=\psi(c_t,x_t)$ and synthesizes a workflow that either creates a new simulator state or edits an existing one. Formally,
\begin{equation}
\begin{aligned}
\pi_t^{\mathrm{env}} &= \phi(z_t^{I},o_t^{S}),\\
\tilde{o}_t^{S} &\xrightarrow{\pi_t^{\mathrm{env}}} o_{t+1}^{S},\\
d\!\left(o_{t+1}^{S},\mathbf{o}^{*}_{\mathrm{op}}(z_t^{I})\right) &= \Delta_{\mathrm{goal}}+\rho\,\Delta_{\mathrm{pres}},
\end{aligned}
\end{equation}
where $\tilde{o}_t^{S}$ denotes either an initialization state or an editable simulator state, $\Delta_{\mathrm{goal}}$ measures consistency with the explicit requirements specified in the user intent, including objects, spatial relations, robot or manipulator configurations, and other environment-level constraints, and $\Delta_{\mathrm{pres}}$ measures unintended changes to content that should remain unchanged during controllable editing.

\paragraph{Trajectory Synthesis.}
EmbodiedClaw models trajectory synthesis as the generation of training-ready data from executable simulator states. This class includes both training trajectory collection and training data transformation, such as converting raw rollouts into model-consumable supervision formats. Given a simulator state $o_t^{S}$ and task specification $x_t$, the system executes a data-generation workflow to produce processed trajectory data $o_t^{D}$. Formally,
\begin{equation}
\begin{aligned}
\pi_t^{\mathrm{traj}} &= \phi(o_t^{S},x_t),\\
(o_t^{S},x_t) &\xrightarrow{\pi_t^{\mathrm{traj}}} o_t^{D},\\
d\!\left(o_t^{D},\mathbf{o}^{*}_{\mathrm{op}}(x_t,o_t^{S})\right) &= \Delta_{\mathrm{task}}+\eta\,\Delta_{\mathrm{stab}}+\kappa\,\Delta_{\mathrm{fmt}},
\end{aligned}
\end{equation}
where $o_t^{D}$ may contain collected trajectories, relabeled transitions, and transformed training samples, $\Delta_{\mathrm{task}}$ measures how well the synthesized data supports the target task, $\Delta_{\mathrm{stab}}$ measures consistency and robustness across repeated rollouts or processing runs, and $\Delta_{\mathrm{fmt}}$ measures fidelity to the desired training-data format.

\paragraph{Model Engineering Operations.}
EmbodiedClaw models model engineering as a unified workflow over model code, weights, and datasets. This class includes automatic code generation and editing, weight and parameter management, automated training given model code, weights, and datasets, and evaluation on one or more target datasets. Given simulator state $o_t^{S}$, dataset state $o_t^{D}$, and model state $o_t^{M}$, the system instantiates an executable pipeline that produces updated model artifacts and evaluation outcomes. Formally,
\begin{equation}
\begin{aligned}
\pi_t^{\mathrm{mdl}} &= \phi(o_t^{S},o_t^{D},o_t^{M}),\\
(o_t^{S},o_t^{D},o_t^{M}) &\xrightarrow{\pi_t^{\mathrm{mdl}}} (o_{t+1}^{M},r),\\
d\!\left((o_{t+1}^{M},r),(o_M^{*},r^{*})\right) &= \Delta_{\mathrm{code}}+\mu\,\Delta_{\mathrm{train}}+\nu\,\Delta_{\mathrm{eval}}+\xi\,\Delta_{\mathrm{res}},
\end{aligned}
\end{equation}
where $o_{t+1}^{M}$ denotes the updated model state, including edited code, managed weights, and trained parameters; $r$ denotes the resulting evaluation outcome; $\Delta_{\mathrm{code}}$ measures the correctness of generated or modified model code and parameter management; $\Delta_{\mathrm{train}}$ measures training effectiveness under the provided code, weights, and datasets; $\Delta_{\mathrm{eval}}$ measures the correctness and completeness of evaluation across target datasets; and $\Delta_{\mathrm{res}}$ measures the associated computational resource usage.

%% file: sections/method.tex
\section{Method}
Building on the formulation in Sec.~2, we instantiate EmbodiedClaw as a conversational execution system for embodied AI development. At each turn $t$, the system takes user input $x_t$ together with the current operational context $\mathbf{o}_t=(o_t^S,o_t^D,o_t^M)$, infers an intent representation, and executes a workflow that updates simulation, data, and model states in a closed loop. The central design principle is to decouple \emph{intent inference}, \emph{workflow planning}, and \emph{platform execution}, so that the same high-level user request can be reliably realized across heterogeneous simulators, assets, datasets, and model stacks.

\begin{figure}[t]
    \centering
    \includegraphics[width=\linewidth]{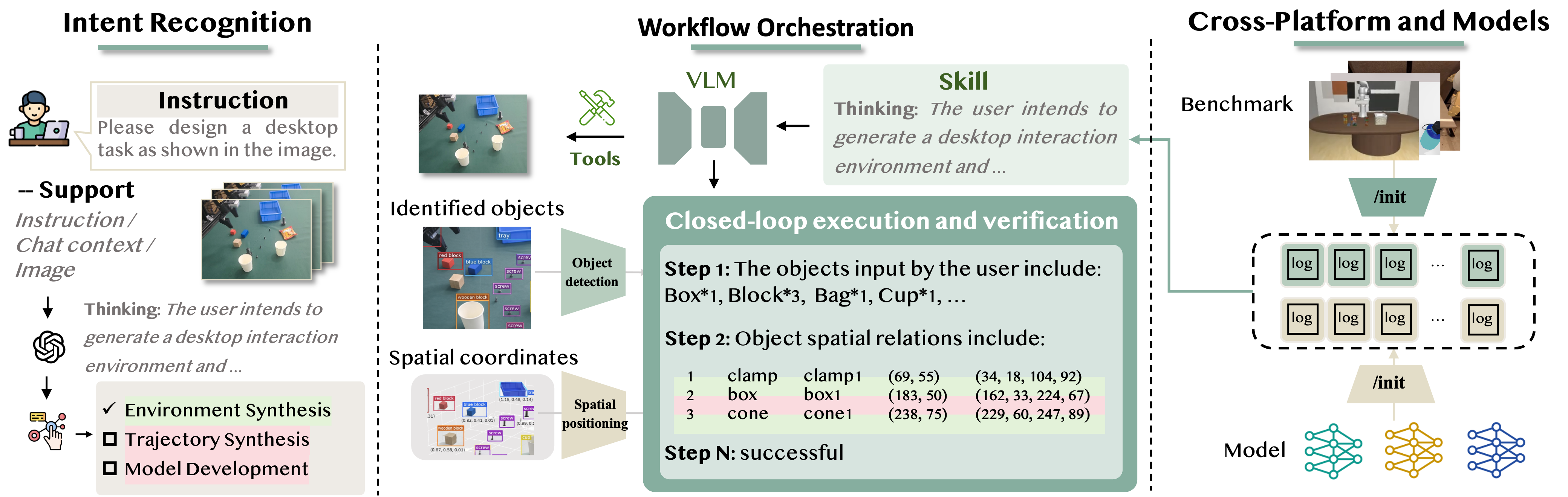}
    \caption{\textbf{EmbodiedClaw pipeline.} The system maps user requests to intent-aware workflows, grounds abstract skills to platform-specific actions, and verifies each step in a closed loop.}
    \label{fig:embodiedclaw_framework}
\end{figure}

\subsection{Framework Overview}
EmbodiedClaw is organized into four modules: intent understanding, workflow orchestration, skill-grounded execution, and asset-platform adaptation, together with a verifier that operates after each execution step. At turn $t$, the end-to-end computation is written as
\begin{equation}
(z_t^{I},\mathcal{O}_t)=\phi_{\mathrm{IU}}(x_t,\mathbf{o}_t),
\qquad
\pi_t=(g_1,\ldots,g_K)=\phi_{\mathrm{WO}}(z_t^{I},\mathcal{O}_t,\mathbf{o}_t),
\end{equation}
\begin{equation}
\mathbf{o}_t^{k+1}=\phi_{\mathrm{AP}}\big(g_k,\mathbf{o}_t^k,\mathcal{A}_t,\mathcal{P}_t\big),
\qquad
v_t^k=\phi_{\mathrm{V}}\big(g_k,\mathbf{o}_t^k,z_t^{I},\mathcal{O}_t\big),
\qquad
\mathbf{o}_{t+1}=\mathbf{o}_t^{K+1},
\end{equation}
where $\mathbf{o}_t^1=\mathbf{o}_t$. Here, $z_t^{I}$ denotes the inferred intent representation, and $\mathcal{O}_t$ denotes the target operational objects, $\mathcal{A}_t$ denotes the active asset set, and $\mathcal{P}_t$ denotes the simulator or platform backend. This decomposition separates semantic decision making from low-level execution, allowing EmbodiedClaw to preserve a stable conversational interface while adapting to diverse embodied development environments.

\subsection{Intent Understanding}
The intent understanding module predicts \emph{what} operation should be performed and \emph{which} operational objects should be updated. Given $(x_t,\mathbf{o}_t)$, it outputs $(z_t^{I},\mathcal{O}_t)$, where $z_t^{I}\in\mathcal{Z}^{I}$ and $\mathcal{O}_t\subseteq\{o_t^S,o_t^D,o_t^M\}$. The input $x_t$ may include conversational context, explicit task instructions, and multimodal signals such as images or other structured observations. In contrast to treating raw user input as directly executable, this module converts under-specified conversational requests into a structured representation that can support downstream planning and verification.

In practice, the intent space $\mathcal{Z}^{I}$ covers the high-frequency operations introduced in Sec.~2, including simulation environment synthesis, trajectory synthesis, and model engineering. The module also performs object grounding, which resolves user mentions into executable references such as scene identifiers, robot configurations, dataset versions, checkpoints, or code assets. This grounding step is critical because downstream execution must operate on concrete entities rather than ambiguous spans in natural language.

\subsection{Workflow Orchestration and Skill-Grounded Execution}
Given $(z_t^{I},\mathcal{O}_t,\mathbf{o}_t)$, the orchestrator generates a skill sequence $\pi_t=(g_1,\ldots,g_K)$ from a reusable skill library $\mathcal{G}$. Each skill exposes a normalized interface over the operational objects,
\begin{equation}
g_k:\ (o^S,o^D,o^M) \mapsto (o'^S,o'^D,o'^M),
\end{equation}
with explicit input-output contracts that support composition across heterogeneous workflows. At this level, skills remain abstract: they encode \emph{what} transformation should occur, but not yet \emph{how} it is realized on a particular backend.

The skill library spans both perception-oriented and action-oriented embodied development primitives. Representative skills include object recognition, spatial localization, asset matching and retrieval, simulator-state editing, trajectory generation, dataset transformation, code editing, training launch, and evaluation dispatch. This design allows the orchestrator to compose heterogeneous capabilities---for example, using object recognition and spatial localization to ground a user request to scene entities, followed by asset matching to retrieve compatible resources, and then invoking execution skills to modify the simulator, generate data, or update model artifacts.

Execution then grounds each abstract skill into concrete operations such as API calls, script invocations, simulator controls, dataset transformations, or model code updates. This separation between \emph{workflow semantics} and \emph{platform realization} is central to EmbodiedClaw. It allows the same workflow pattern to be reused across distinct simulators, asset libraries, and model stacks, while confining platform-specific complexity to the grounding layer.

\subsection{Closed-Loop Verification and Recovery}
EmbodiedClaw performs step-wise verification after every skill call to improve reliability over long-horizon, multi-stage workflows. The verifier outputs
\begin{equation}
v_t^k=(\texttt{status},\texttt{message},\texttt{rollback}),
\end{equation}
where \texttt{status} indicates whether the current step succeeds, \texttt{message} records structured diagnostics, and \texttt{rollback} indicates whether the latest valid state should be restored.

When verification fails, the system halts unsafe downstream execution, rolls back if necessary, and feeds the diagnostic signal back to the orchestrator for recovery, such as argument repair, skill substitution, or workflow replanning. This closed-loop design reduces error accumulation, improves state consistency across long execution chains, and enables EmbodiedClaw to recover from execution-time failures rather than treating them as terminal outcomes.

\subsection{Asset and Platform Adaptation}
EmbodiedClaw treats assets and simulator backends as pluggable execution contexts rather than as part of the workflow definition itself. Given an abstract workflow $\pi_t$, the adaptation module binds object references to concrete assets $\mathcal{A}_t$ and backend $\mathcal{P}_t$,
\begin{equation}
\phi_{\mathrm{AP}}(\pi_t,\mathcal{A}_t,\mathcal{P}_t).
\end{equation}
Beyond binding existing assets, EmbodiedClaw maintains a predefined asset library for commonly used embodied objects and environments, while also supporting optional third-party 3D asset sources for long-tail asset acquisition. To bridge these sources with executable simulators, we implement dedicated skills for asset retrieval, downloading, normalization, and automatic registration into the target simulation environment. As a result, a user request can be resolved either to an existing in-library asset or to a newly downloaded external asset that is automatically imported and registered for subsequent workflow execution.

This design enables a workflow to be transferred across platforms without rewriting the user-facing interaction logic or the high-level plan structure, while also expanding the reachable asset space beyond the predefined library. In our implementation, EmbodiedClaw is instantiated on RoboTwin, where the adaptation layer supports asset registration and expansion, third-party asset ingestion, and cross-platform execution under a unified conversational interface.

%% file: sections/experiments.tex
\section{Experiments}

EmbodiedClaw is designed to address three central challenges in embodied AI development: reducing the engineering effort required to build and iterate embodied tasks, improving the correctness of executable development outcomes, and maintaining robustness under complex multi-step workflows. To evaluate the capabilities of EmbodiedClaw, we construct an experimental suite on RoboTwin that covers the main stages of embodied development, including intent-to-simulation construction, conversational environment editing, trajectory collection, model training and evaluation, and long-horizon workflow execution.

Based on this experimental setup, our evaluation focuses on the following three key questions:

\begin{list}{\arabic{enumi}.}{\usecounter{enumi}\setlength{\leftmargin}{2.2em}\setlength{\itemsep}{0pt}\setlength{\parsep}{0pt}\setlength{\topsep}{2pt}}
    \item Can EmbodiedClaw significantly improve the efficiency of embodied AI development?
    \item Can EmbodiedClaw improve the accuracy and executability of embodied development tasks?
    \item What can EmbodiedClaw do across embodied AI development workflows?
\end{list}

\subsection{Can EmbodiedClaw Improve the Efficiency of Embodied AI Development?}
\begin{figure}[htb]
    \centering
    \includegraphics[width=\linewidth]{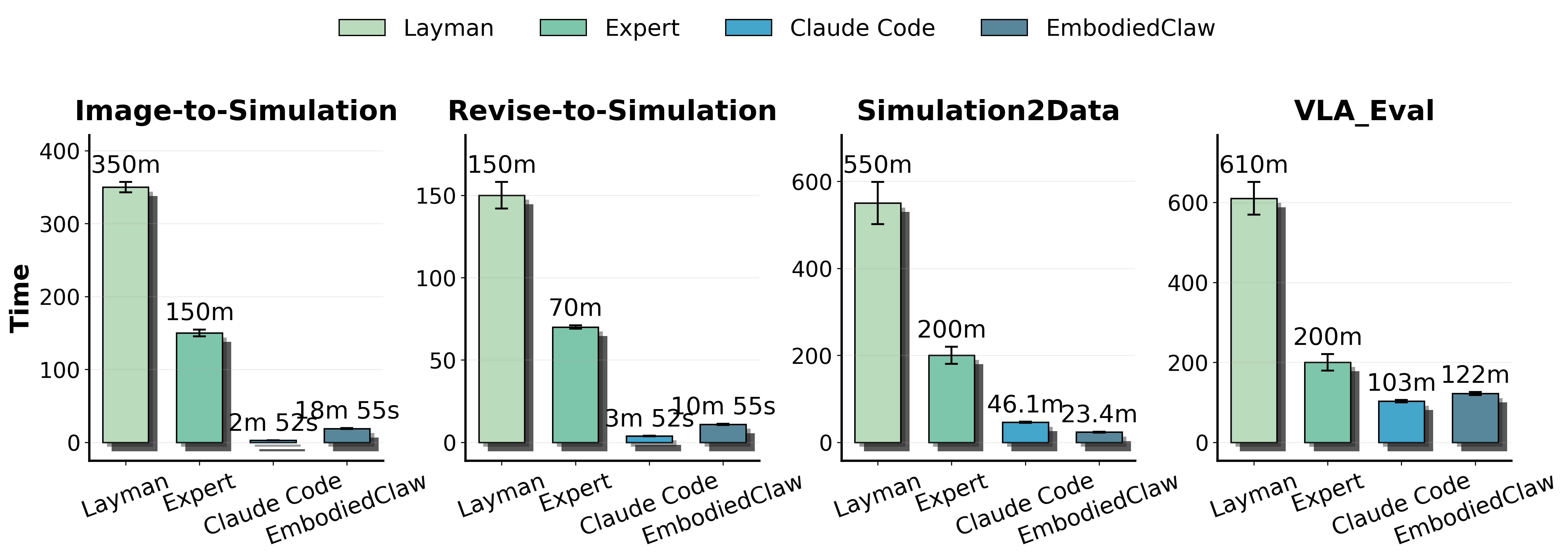}
    \caption{\textbf{Efficiency comparison across four embodied AI development tasks.} The four subfigures report the efficiency of \textbf{Layperson}, \textbf{Expert}, \textbf{Claude Code}, and \textbf{EmbodiedClaw} on \emph{image-to-simulation}, \emph{revise-to-simulation}, \emph{simulation-to-data}, and \emph{VLA evaluation}, respectively. Each subfigure compares the time cost of completing the corresponding task, where lower is better.}
    \label{fig:efficiency_comparison}
\end{figure}

We evaluate four representative tasks: (1) \textbf{Environment Synthesis}, which creates 10 simulation environments from reference images; (2) \textbf{Environment Synthesis}, which revises these 10 environments according to user-specified modification requests; (3) \textbf{Trajectory Synthesis}, which collects trajectories from these 10 environments and converts them into training-ready data formats; and (4) \textbf{Model Engineering Operations}, which evaluates two VLA models, \textsc{ACT} and \textsc{RDT}, on the RoboTwin dataset. We compare four settings: \textbf{Layperson}, \textbf{Expert}, \textbf{Claude Code}, and \textbf{EmbodiedClaw}, with three participants in each setting. For the human-user settings, the \textbf{Expert} group is composed of Ph.D. students actively engaged in embodied intelligence research, whereas the \textbf{Layperson} group is composed of M.S. and Ph.D. students in computer science and related fields who are technically trained but lack prior experience with embodied benchmarks, simulator configuration, and platform-specific workflows. We use average task completion time as the efficiency metric, where lower is better.

As shown in Fig.~\ref{fig:efficiency_comparison}, EmbodiedClaw substantially improves efficiency over both \textbf{Layperson} and \textbf{Expert} across all four tasks. The improvement is particularly pronounced on high-cost and complex workflows. For example, on \emph{simulation-to-data}, EmbodiedClaw reduces completion time from 200 minutes to 23.4 minutes relative to \textbf{Expert}, corresponding to an efficiency improvement of approximately 88.3\%. On \emph{VLA evaluation}, it reduces completion time from 200 minutes to 122 minutes, yielding a 39.0\% efficiency improvement over \textbf{Expert}. These results indicate that EmbodiedClaw is especially effective on tasks whose manual execution requires repeated configuration, tool chaining, and platform-specific operations.

Compared with \textbf{Claude Code}, EmbodiedClaw incurs slightly longer runtime on some tasks because it invokes encapsulated embodied tools and standardized task workflows rather than relying solely on unconstrained code generation. However, this additional runtime leads to substantial gains in execution accuracy and reliability, as shown in Sec.~4.2. Overall, these results show that EmbodiedClaw substantially reduces embodied engineering overhead while delivering a stronger efficiency--accuracy trade-off for executable embodied development.

\subsection{Can EmbodiedClaw Improve Accuracy and Executability?}

\begin{figure}[htb]
    \centering
    \includegraphics[width=\linewidth]{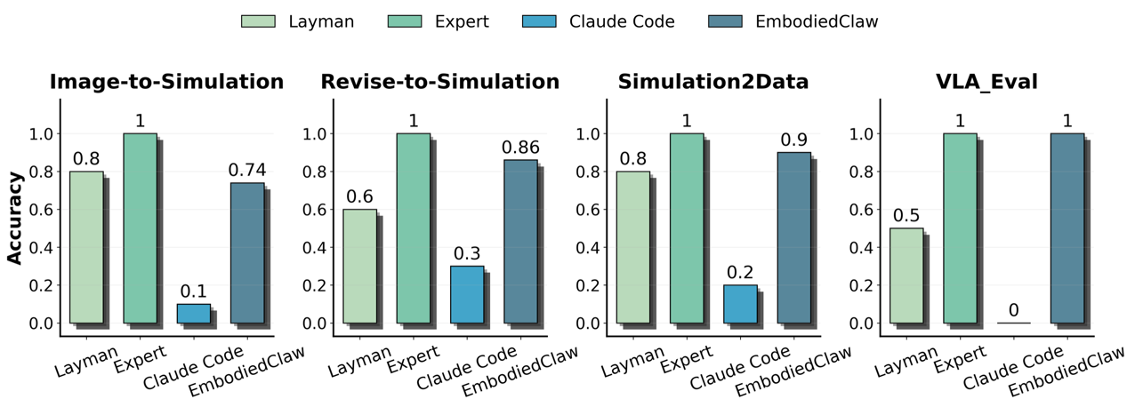}
    \caption{\textbf{Task completion rate comparison across four embodied AI development tasks.} The four subfigures report the accuracy of \textbf{Layperson}, \textbf{Expert}, \textbf{Claude Code}, and \textbf{EmbodiedClaw} on \emph{image-to-simulation}, \emph{revise-to-simulation}, \emph{simulation-to-data}, and \emph{VLA evaluation}, respectively. Higher is better.}
    \label{fig:accuracy_comparison}
\end{figure}
We further evaluate EmbodiedClaw in terms of task completion rate, which measures whether an embodied development workflow is completed correctly and executed successfully. As shown in Fig.~\ref{fig:accuracy_comparison}, EmbodiedClaw consistently achieves performance closest to human experts across the four evaluated tasks. This result shows that EmbodiedClaw not only improves development efficiency, but also produces reliable and executable outcomes across diverse embodied development settings.

The advantage is especially clear on complex tasks with stronger execution dependency, such as trajectory synthesis and VLA evaluation. On \emph{simulation-to-data}, EmbodiedClaw attains an accuracy of 0.9, approaching expert-level performance of 1.0 and substantially outperforming \textbf{Claude Code} (0.2). On \emph{VLA evaluation}, EmbodiedClaw reaches an accuracy of 1.0, matching human experts and significantly exceeding \textbf{Layperson} (0.5) and \textbf{Claude Code} (0.0). These results indicate that EmbodiedClaw is particularly effective on tasks that require correct execution across multiple dependent stages.

Compared with \textbf{Claude Code}, EmbodiedClaw performs markedly better across all four tasks, highlighting a central point of our work: strong general-purpose agent capability alone is insufficient for complex embodied development. In these settings, success depends not only on generating plausible code or actions, but also on correctly grounding user intent, decomposing tasks into executable stages, invoking the right tools at each stage, and maintaining consistency across simulator, data, and model states. EmbodiedClaw is designed precisely around this requirement. Through strict workflow orchestration and reliable tool-grounded execution, it converts under-specified embodied requests into standardized, verifiable procedures, which substantially improves end-to-end executability and correctness.

\subsection{What Can EmbodiedClaw Do Across Embodied AI Development Workflows?}

\begin{figure}[htb]
    \centering
    \includegraphics[width=\linewidth]{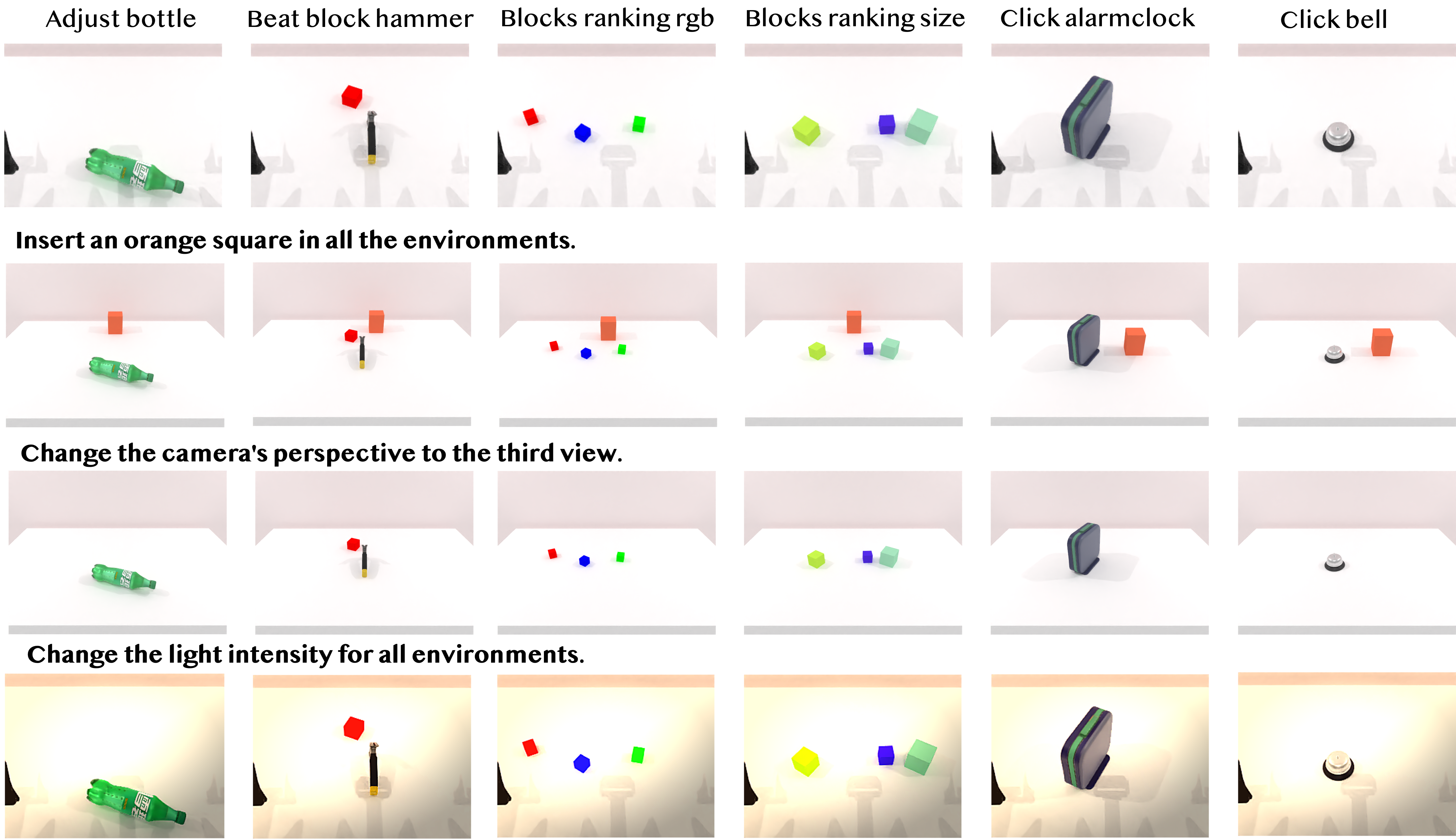}
    \caption{\textbf{Case study of batch editing for existing environments.} EmbodiedClaw supports parallel editing of existing environments from natural-language instructions, including inserting new objects, modifying irrelevant viewpoints, and adjusting scene lighting conditions.}
    \label{fig:case_example1}
\end{figure}

\begin{figure}[htb]
    \centering
    \includegraphics[width=\linewidth]{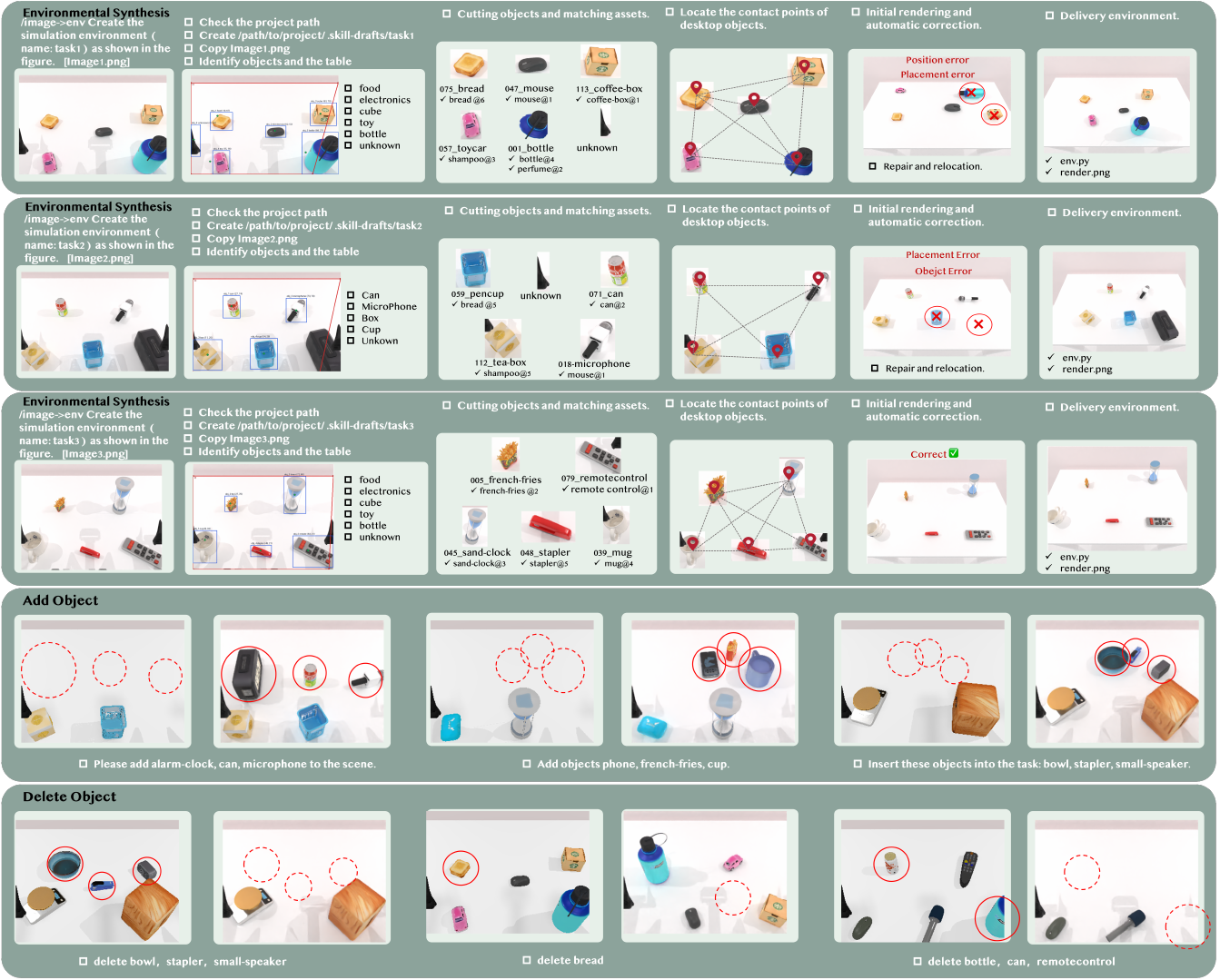}
    \caption{\textbf{Case studies of EmbodiedClaw for environment construction and editing.} Given natural-language instructions, EmbodiedClaw translates user requests into executable workflows and completes the corresponding environment construction and editing procedures automatically.}
    \label{fig:case_example2}
\end{figure}

\begin{figure}[htb]
    \centering
    \includegraphics[width=\linewidth]{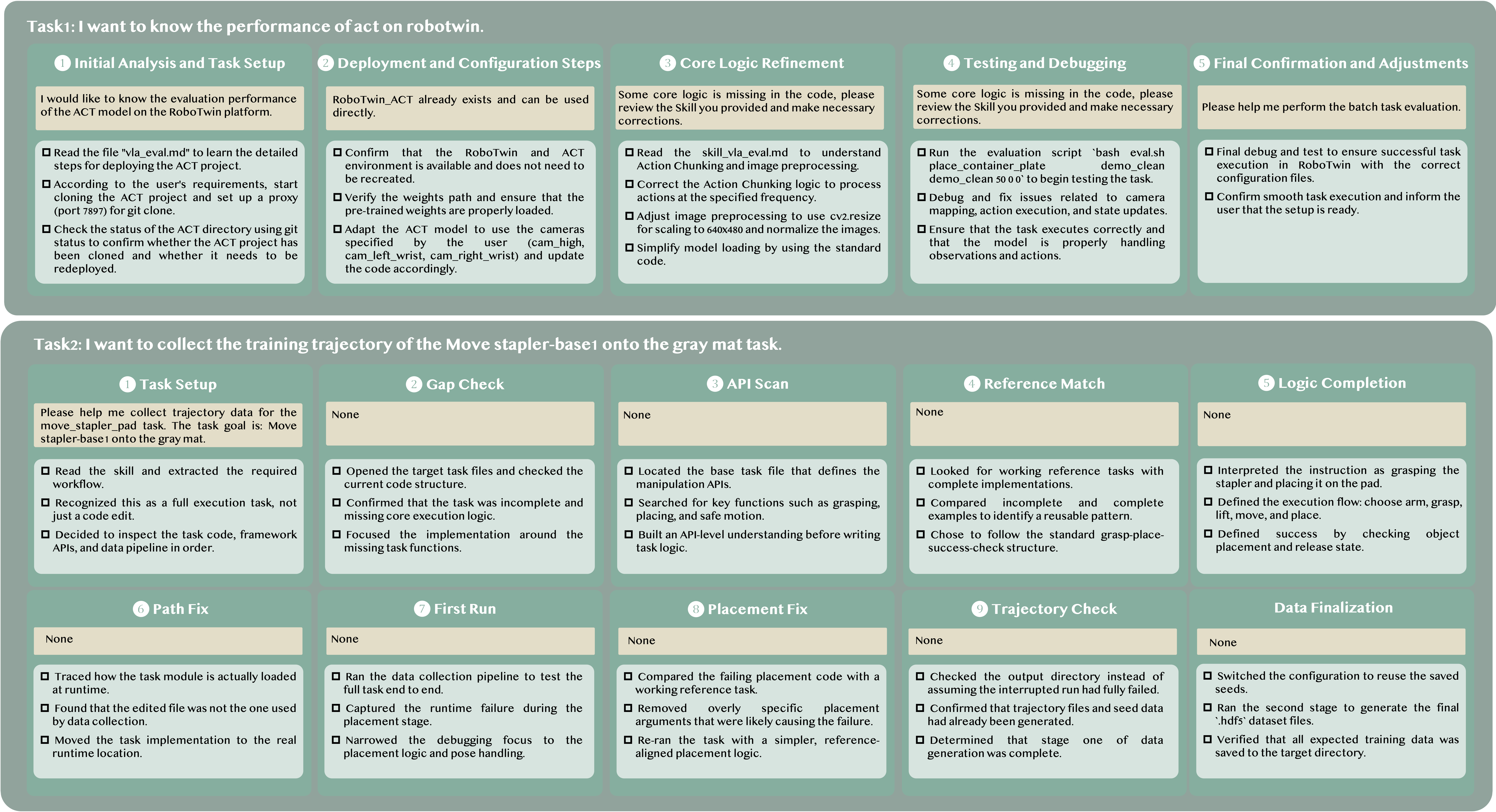}
    \caption{\textbf{Case study of the VLA evaluation workflow.} EmbodiedClaw executes VLA evaluation as a complete workflow, covering environment preparation, trajectory collection, data transformation, model deployment, and downstream evaluation.}
    \label{fig:case_example3}
\end{figure}

The three case studies in Fig.~\ref{fig:case_example1} further demonstrate that EmbodiedClaw improves executability across embodied AI development workflows. Across environment construction, scene editing, data generation, and downstream evaluation, EmbodiedClaw can consistently turn user intent into executable operations rather than incomplete intermediate outputs. This shows that its benefit is not limited to a single task type, but extends to diverse workflows that require correct execution under realistic embodied settings.

More importantly, these examples in Fig.~\ref{fig:case_example2} and ~\ref{fig:case_example3} help explain why EmbodiedClaw achieves higher accuracy and executability. By converting under-specified natural-language requests into structured procedures, invoking the appropriate tools, and preserving consistency across environment, data, and model states, EmbodiedClaw reduces failures caused by missing steps, incorrect tool usage, and broken workflow transitions. As a result, it is better able to complete embodied development tasks correctly and reliably, especially when execution depends on multi-stage coordination.

%% file: sections/related_works.tex

\section{Related Work}
\label{sec:related_work}

\paragraph{Repetitive and Time-Consuming Operations in Embodied AI Research.}
Embodied AI has witnessed explosive growth in simulation-based research,
with a wide range of platforms---including AI2-THOR~\cite{kolve2017ai2thor},
Habitat~\cite{savva2019habitat,szot2021habitat2},
ManiSkill3~\cite{tao2024maniskill3}, Isaac Gym~\cite{makoviychuk2021isaac},
SAPIEN~\cite{xiang2020sapien}, and VirtualHome~\cite{puig2018virtualhome}---each
providing distinct interfaces, asset formats, and physics engines.
Despite this diversity, constructing even a single research pipeline spanning
data generation, simulation configuration, policy training, and evaluation
demands substantial manual effort across all of these platforms.
As highlighted by several recent works~\cite{duan2024survey,ma2025ecp,firoozi2025foundation,zhang2025generative},
a typical embodied AI workflow stitches together these stages through
proprietary interfaces and task-specific integration logic, imposing heavy
recurring overhead on every new research iteration.
Furthermore, the absence of unified interfaces means that transitions between
perception, planning, and control modules still rely on manually configured
logic, limiting reusability and increasing integration overhead across
deployment scenarios~\cite{ma2025ecp,liu2023survey}.
These repetitive, time-consuming operations remain a persistent bottleneck
that fundamentally limits research throughput in the field.

\paragraph{Automation in Embodied AI Research.}
Several efforts have begun to automate isolated stages of the embodied AI
research pipeline.
For environment construction, ProcTHOR~\cite{deitke2022procthor} procedurally
generates large-scale interactive indoor environments, and
Holodeck~\cite{yang2024holodeck} uses LLMs to automatically build diverse 3D
scenes from language prompts.
More recently, SAGE~\cite{xia2026sage} proposes an agentic framework that
generates simulation-ready 3D environments through iterative reasoning and
adaptive tool selection, while EmbodiedGen~\cite{wang2025embodiedgen} provides
a generative platform for producing physics-grounded 3D assets directly
importable into mainstream simulators such as MuJoCo~\cite{todorov2012mujoco}
and Isaac Sim.
For skill and data acquisition, RoboGen~\cite{wang2023robogen} decomposes
high-level tasks into sub-tasks, selects an appropriate learning
paradigm---reinforcement learning, motion planning, or trajectory
optimization---and generates training supervision automatically.
M$^2$Diffuser~\cite{m2diffuser2024} similarly constructs automated data
collection pipelines for mobile manipulation trajectories.
Scenethesis~\cite{ling2025scenethesis} further integrates LLM-based scene
planning with vision-guided layout refinement to generate physically plausible
interactive environments.
Despite these advances, existing work remains \textbf{fragmented and mutually
incompatible}: each system targets a single pipeline stage, exposes distinct
interfaces, and operates on incompatible data formats.
Even within individual frameworks, stage transitions still rely on manually
configured logic, limiting reusability and increasing integration
overhead~\cite{ma2025ecp}.
No unified system addresses the full embodied AI research workflow end-to-end.

\paragraph{AI-Powered Workflow Automation Systems.}
General-purpose AI agent frameworks have demonstrated strong automation
capability across broad domains.
Systems such as OpenClaw~\cite{openclaw2026, zhao2026cutclaw} accept natural-language commands
and autonomously execute complex multi-step workflows---from web scraping and
code submission to file management---serving as versatile personal automation
agents.
Their key strength lies in flexible, LLM-driven orchestration: the agent
evaluates available tools and chains them into execution pipelines through a
unified workflow shell~\cite{openclaw2026}.
Related agentic frameworks, including those built on
LangChain~\cite{liu2023survey}, AutoGen~\cite{wu2024autogen}, and CrewAI, further support
multi-agent collaboration and dynamic task planning for general-purpose
workflows.
However, such systems exhibit clear limitations when applied to tasks requiring
deep domain expertise~\cite{masterman2024landscape,yang2025survey}.
Applied to embodied AI research---which demands specialized knowledge in
simulator configuration, robot kinematics, cross-platform asset handling, and
training pipeline orchestration---general-purpose agents lack the requisite
domain knowledge to construct valid and reproducible workflows, and thus fail
to deliver meaningful productivity gains.
To bridge this gap, we present \textbf{EmbodiedClaw}, a workflow automation
system purpose-built for embodied AI research.
EmbodiedClaw encapsulates domain knowledge across scene construction, trajectory
collection, training configuration, and cross-platform evaluation into
natural-language-driven automated pipelines, enabling researchers to accomplish
complex tasks simply through chat.

%% file: sections/conclusion.tex
\section{Conclusion}

We presented EmbodiedClaw, a conversational workflow execution system for embodied AI development. By grounding user intent in executable operations over environments, data, and models, EmbodiedClaw transforms high-frequency embodied research activities, including environment synthesis and editing, trajectory collection, model training, and evaluation, into unified and executable workflows. Experiments show that EmbodiedClaw improves development efficiency while also increasing executability and reliability across diverse embodied AI tasks.